\documentclass[final,12pt]{elsarticle}

\usepackage{natbib}
\usepackage[utf8]{inputenc} 
\usepackage[T1]{fontenc}    
\usepackage{hyperref}       
\usepackage{url}            
\usepackage{booktabs}       
\usepackage{amsfonts}       
\usepackage{nicefrac}       
\usepackage{microtype}      
\usepackage{graphicx}
\usepackage{amsmath,amsthm,amssymb}
\usepackage[noend]{algpseudocode}
\usepackage{algorithm}
\usepackage[noend]{algpseudocode}
\usepackage{multirow}
\usepackage{breqn}
\usepackage{subfig}
\usepackage{rotating}
\usepackage{pdflscape}
\usepackage{color}
\usepackage{setspace} 
\usepackage{lineno}
\usepackage[dvipsnames]{xcolor}

\journal{Applied Soft Computing}

\begin{document}
\begin{frontmatter}

\title{Unsupervised Feature Learning for Environmental Sound Classification Using Weighted Cycle-Consistent Generative Adversarial Network}

 \author{Mohammad Esmaeilpour\corref{}}
 \ead{mohammad.esmaeilpour.1@ens.etsmtl.ca} 
 \author{Patrick Cardinal}
 \ead{patrick.cardinal@etsmtl.ca} 
  \author{Alessandro L. Koerich}
 \ead{alessandro.koerich@etsmtl.ca} 
 \address{École de Technologie Supérieure, Université du Québec \\ 1100 Notre-Dame West, Montréal, QC, H3C 1K3, Canada.}

\begin{abstract}
In this paper we propose a novel environmental sound classification approach incorporating unsupervised feature learning via the spherical $K$-Means++ algorithm and a new architecture for high-level data augmentation. The audio signal is transformed into a 2D representation using a discrete wavelet transform (DWT). The DWT spectrograms are then augmented by a novel architecture for cycle-consistent generative adversarial network. This high-level augmentation bootstraps generated spectrograms in both intra- and inter-class manners by translating structural features from sample to sample. A codebook is built by coding the DWT spectrograms with the speeded-up robust feature detector and the $K$-Means++ algorithm. The Random forest is the final learning algorithm which learns the environmental sound classification task from the code vectors. Experimental results in four benchmarking environmental sound datasets (ESC-10, ESC-50, UrbanSound8k, and DCASE-2017) have shown that the proposed classification approach outperforms most of the state-of-the-art classifiers, including convolutional neural networks such as AlexNet and GoogLeNet, improving the classification rate between 3.51\% and 14.34\%, depending on the dataset. 
\end{abstract}

\end{frontmatter}

\section{Introduction}
Environmental sound classification has been attracted the interest of several researchers in machine learning because of its vast applications~\cite{chu2009environmental, radhakrishnan2005audio, xu2008audio}. However, this is a challenging problem due to the complex nature of environmental sounds in terms of dimensionality, different mechanism of sound production, overlapping of different sources, and lack of high-level structures usually observed in speech and in many types of musical sounds~\cite{salamon2015unsupervised}. This complex nature masked by natural acoustic noises~\cite{noda1987effect} can make the classification of specific sounds very challenging. This challenge becomes more difficult when audio classes do not have similar sound production mechanisms such as "car horn" and "car engine idling" in open scenes like streets or parks~\cite{chu2009environmental,ellis2004minimal,chaudhuri2013unsupervised}.

In the literature, the classification of environmental sounds has been addressed using both standalone and ensemble classification setups incorporating conventional classifiers and deep neural networks where the input signal can be represented by an audio waveform (1D) or converted to a mid-level representation (2D) such as a spectrogram~\cite{salamon2015unsupervised, aytar2016soundnet, dai2017very, mun2017generative, piczak2015environmental, salamon2017deep, tokozume2017learning}. The audio signal may also be represented by handcrafted features in the spectral or the cepstral domains mainly via frequency transformations which are lossy operations. Zero-crossing rate~\cite{lu2002content}, spectral flux and centroid~\cite{tzanetakis2002musical}, chroma vector~\cite{ellis2007classifying}, Mel frequency cepstral coefficients (MFCCs)~\cite{logan2000mel}, short-time Fourier transform (STFT)~\cite{smith2011spectral}, cross recurrent plot (CRP)~\cite{serra2009cross}, and discrete wavelet transform (DWT)~\cite{van2011discrete} are among the most well-known handcrafted features for audio classification~\cite{papakostas2017deep}. These handcrafted features not only reduce the dimensionality of the audio signal but may also reduce some types of noise and help to extract time-varying descriptors which provide a better discrimination. The approaches that use these features have shown relatively better performance than the approaches that use 1D signals directly in both classification and clustering tasks, mainly when employing conventional classifiers such as support vector machines (SVMs)~\cite{gerek2008compression}. 

MFCC is a common and reliable informative representation format for analyzing audio and for this reason, most of the proposed classification approaches in this domain rely on it~\cite{radhakrishnan2005audio,cai2006flexible, ganchev2005comparative, heittola2013context}. MFCCs are handcrafted features based on the human auditory system, which can make a reasonable balance between handling the complex nature of real-life sounds and providing informative feature vectors for classification purposes. In addition to traditional classifiers such as Gaussian mixture models~\cite{godino2006dimensionality}, hidden Markov models~\cite{gales1992improved} and $K$-nearest neighbor~\cite{eronen2006audio}, convolutional neural networks (ConvNets)~\cite{deng2013new} have been evaluated on MFCC feature vectors and achieved better results than the 1D audio signal. However, MFCCs have shown to be very sensitive to background noise and this might affect the performance of classifiers for noisy environmental sounds~\cite{cotton2011spectral}.

With the recent advances in deep learning, many strong classifiers such as ConvNets have been introduced, which are designed to learn directly both from 1D and 2D data. ConvNets are quite similar to dense deep neural network (DNN) where the main difference is the inclusion of convolution layer(s) to deal with raw data. The main advantage of these networks is their ability to learn directly from raw data rather than handcrafted features. ConvNets have been used with audio waveforms with several convolution layers incorporating different 1D signal augmentation methods~\cite{salamon2017deep,abdoli2019}. Experimental results have shown competitive accuracy compared to unsupervised sound classifiers~\cite{salamon2015unsupervised}, ConvNets on MFCCs~\cite{piczak2015esc}, and even better performance~\cite{palaz2015convolutional} depending on the dataset. ConvNets have also been evaluated with a combination of 1D and MFCC feature vectors which resulted in low classification error~\cite{tokozume2017learning}. This shows the importance of the representation space in extracting discriminating features.

Audio signals are high dimensional, which means that more than a thousand floating-point values need to be used to represent a short audio signal. Due to this fact, it is preferred to train classifiers on 2D audio representations over audio waveforms. Although, ConvNets have shown great classification performances in 1D signal format~\cite{salamon2017deep,abdoli2019}, so far, they could not outperform AlexNet and GoogLeNet on STFT, DWT, and CRP spectrograms~\cite{weiping2017acoustic}. The majority of recent papers in audio classification especially environmental sounds are on 2D representations mainly for DNNs such as the networks introduced in~\cite{deng2013new}. STFT, DWT, and CRP are the main approaches for producing spectrograms and they can also be combined to augment the amount of data and to extract more informative 2D representations for training ConvNets \cite{boddapati2017classifying}. It has been shown that STFT and DWT have more competence for extracting temporal and structural content for ConvNets~\cite{wyse2017audio}. For some common environmental datasets, GoogLeNet and AlexNet have achieved the highest recognition accuracy with quite high confidence.   

However, one of the main bottlenecks for using ConvNets in environmental sound classification is the amount of data required to train such networks properly due to the high number of parameters to adjust. The two main approaches that have been used to circumvent this problem are: (i) fine-tuning ConvNets pre-trained on other domains/datasets; (ii) generating artificial samples by data augmentation. Both 1D and 2D data augmentation approaches~\cite{weiping2017acoustic,salamon2015feature} have been proposed for improving classification performance which proves the importance of providing better input rather than implementing highly complex and costly networks ~\cite{mun2017generative}. There are several algorithms for augmenting a dataset both in terms of enhancing samples' visual quality and quantity. Augmentation in 2D representations like spectrograms is mostly being implemented with low-level transformations~\cite{cirecsan2011high} including translation, shearing, rotation, scaling, aspect ratio, flipping, etc., which in general may not improve the performance of conventional or deep learning classifiers. The linear nature of these affine transformations may not cause a high impact on the classifier decision boundaries~\cite{zhu2018emotion}. It is worth mentioning that, even these low-level data augmentations have been sometimes contributed significantly in training ConvNets and reducing overfitting.  

Elastic deformation~\cite{simard2003best} is another type of low-level augmentation which has been used in spectrograms. The elastic deformation implements a similarity transformation which interpolates between highly correlated spectrogram sub-manifolds. However, when the resolution of the spectrogram is small, and it does not have much active areas (super uniform areas in pixel-wise level), this augmentation may not work well especially for deep learning models. Extracting covariant patches and color space channel intensity alteration~\cite{krizhevsky2012imagenet} as well as other types of pixel-level augmentation scheme has been utilized in many spectrogram classification tasks. In addition to the linear nature of low-level augmentation, they cannot enhance data distribution which is usually determined by high-level features. Some methodologies have been proposed for circumventing this problem such as learning multivariate normal distribution for each class with respect to their mean manifolds~\cite{hauberg2016dreaming}. Implementation of this augmentation in the real world, especially for long audio sequences of high dimension is not optimal. One potential solution could be multivariate distribution learning in representation space~\cite{dixit2017aga} with respect to the structural components~\cite{wang2004image} of a spectrogram. 

In this paper, we propose a novel architecture for data augmentation which translates one spectrogram to another using a generative model named Weighted Cycle-Consistent Generative Adversarial Network (WCCGAN), as well as a novel approach for environmental sound classification based on unsupervised feature learning. The proposed approach has four main steps: (i) audio dimension conversion and preprocessing (from 1D to 2D); (ii) data augmentation using the proposed WCCGAN; (iii) extracting feature vectors from the augmented dataset via speeded-up robust feature detector (SURF) algorithm and leaning a codebook of representative codewords; and (iv) training a random forest algorithm on code vectors. The experimental results have shown that our approach outperforms cutting-edge classifiers such as AlexNet and GoogLeNet in four benchmarking environmental sound datasets: ESC-10 \cite{piczak2015esc}, ESC-50 \cite{piczak2015esc}, UrbanSound8k \cite{Salamon:UrbanSound:ACMMM:14} and DCASE-2017 \cite{mesaros2017dcase}. Besides that, the experimental results have also shown the remarkable performance of the proposed data augmentation approach for both the unsupervised feature learning and supervised approaches.

The organization of this paper is as follows. In Section \ref{sec:prepro} we discuss the transformation of audio waveforms (1D) into spectrograms (2D), as well as the preprocessing steps preceding and succeeding such a transformation for data augmentation purposes. Section \ref{sec:CCGAN} presents the WCCGAN for high-level spectrogram augmentation. In Section \ref{sec:featlearn}, we explain our feature learning methodology using SURF descriptors and the spherical $K$-Means++ algorithm, and also the classification approach based on random forests. Section \ref{sec:exp} provides details about the architecture of the proposed WCCGAN and the experiments carried out in four benchmarking datasets. In Section \ref{sec:disc} we compare the importance of pitch-shifting as one of the basic data augmentation approaches over all other algorithms presented in~\cite{salamon2017deep}. We show that, implementing all types of data augmentations does not necessarily produce informative features favorable to the proposed classifier. We also compare the performance of the proposed WCCGAN with the cycle-consistent GAN proposed by Zhu et al.~\cite{zhu2018emotion} to emphasize the importance of adapting generative model architectures according to the application. The conclusions and future work are presented in the last section.

\section{Preprocessing and Spectrogram Generation}
\label{sec:prepro}
In this section we present the preprocessing steps to artificially expand the size of an audio dataset by creating modified versions of the audio clips and the strategy used to convert such audio clips into spectrograms.  

\subsection{1D Data Augmentation}
Given the relatively small size of the environmental sound datasets, one of the recommended steps before transforming an audio waveform to a 2D representation is to boost the amount, distribution, and cardinality of the samples of each class in the datasets. Data augmentation can be carried out by applying some filters on an audio signal such as pitch shifting, time stretching, compressing dynamic range, and background noise removal~\cite{salamon2017deep}. These operations can be individually applied to an audio sample to produce slightly modified versions of it and increase the number of samples. Finally, these crafted samples are added to the original dataset. Since these augmentation filters increase the number of samples of the dataset, they may have potential to affect the performance of data-driven classifiers.

It has been shown~\cite{boddapati2017classifying} that the pitch-shifting filter alone can highly boost the quality of audio recordings when compared with applying all above-mentioned augmentation filters as proposed in~\cite{salamon2017deep}. After conducting several exploratory experiments, we have found out that, for most of the environmental sound datasets applying all 1D data augmentation filters do not necessarily produce good audio samples in terms of producing samples with low inter-class and high intra-class similarity. In Section \ref{sec:disc} we show some experimental results that support this claim. Therefore, we only use the pitch-shifting augmentation as its constructive effects have been shown in~\cite{salamon2017deep} for both supervised and unsupervised feature learning. For such an aim, we use static pitch shifting scales~\cite{mcfee2015software}. This boosts the number of samples in the dataset with respect to the number of applied scales. 

\subsection{Spectrogram Generation}
STFT, DWT, and CRP are the main approaches for producing spectrograms for an audio signal. ConvNets have shown strong capability in learning from these spectrograms either standalone~\cite{weiping2017acoustic} or pooled together~\cite{boddapati2017classifying}. The DWT representation is more stable to time warping deformations and it can better characterize time varying structures compared to other representations such as STFT~\cite{mallat2012group}. The STFT transformation is somewhat similar to DWT in terms of producing low and high frequency components encoded as spectrograms. Considering a discrete-time signal $x[n]$, its DWT transformation is given by:
\begin{equation}
\mathrm{DWT}(x[n])=(x[n] \otimes g[n]) = \sum \limits _{{k=-\infty }}^{\infty }{x[k]g[n-k]}
\end{equation}
\noindent where $\otimes$ denotes the convolution of $x[n]$ and $g[.]$ (mother function which produces other signals which can be either low or high pass filter sets.) This operation can be applied to at most the minimum length of $x[n]$. The 2D representation of this signal can be computed by:
\begin{equation}
\mathrm 
{S_{DWT}} \{x[n]\}\equiv |\mathrm{DWT}(x[n])|^{2}
\end{equation}
\noindent where $S_{DWT}$ denotes the spectrogram of the signal $x[n]$.

We generate DWT spectrograms using our modified version of the sound explorer C++ script~\cite{waveletSoundExplorer} for the original audio signals as well as the pitch-shifted audio samples to handle audio clips with any length (time duration).

\subsection{Spectrogram Enhancement}
Each generated spectrogram is a 2D array of intensity values which can be noisy when its associated audio signal is affected by environmental noise(s). In this case, adjusting the distribution of the intensity values can help to extract/learn more informative features~\cite{segura2002feature}. For improving the color space and the dynamic color contrast of the intensity values, we apply a histogram equalization filter~\cite{gonzalez2002digital}. Considering each pixel intensity of the generated spectrogram $S$ as $S(i,j)$, then the enhanced spectrogram ($S_{heq}$) is defined in Equation \ref{histogram_equal}. 

\begin{equation}
S_{heq}(i,j)=\left \lfloor (s-1)\sum_{i=0}^{S(i,j)} p_{i}\right \rfloor
\label{histogram_equal}
\end{equation}

\noindent where $s$ is the supremum of 8-bit precision and $p_{i}$ denotes the ratio of pixels with intensity $i$ over the total number of pixels. This filter expands the intensity range of a given spectrogram in a balanced distribution. In the next section we explain how to structurally augment the generated spectrograms towards more informative samples.

\section{Weighted Cycle-Consistent Generative Adversarial Network (WCCGAN)}
\label{sec:CCGAN}
The approach proposed for augmenting generated spectrograms is based on the Cycle-Consistent Generative Adversarial Network (CCGAN) which maps one spectrogram to another spectrogram. The efficiency of this 2D-to-2D translation has been proven in the literature to image-to-image translation problems~\cite{isola2017image,zhu2017unpaired}. The proposed GAN architecture is inspired by Zhu et al.~\cite{zhu2017data} with two main differences: (i) it incorporates two identity mapping functions for avoiding the oversmoothing of generated spectrograms, which affects the performance of the discriminator towards a wrong label other than the pre-defined target label; (ii) it employs different architectures for both generator and discriminator.

The proposed augmentation pipeline is implemented only in 2D space since mapping 1D-to-1D audio signal for augmentation purposes is very challenging due to the high dimensionality of audio signals. Our perspective in data augmentation is directed towards increasing inter- and intra-class structural contents over low-level pixel augmentations. This can help classifiers to reach a finer decision boundary among data sub-manifolds with minimum overlap. A more accurate way to impose structural contents on data augmentation is by using GAN since we can consistently control the mapping process from one image to another by adding an extra constraint to its loss function(s).

\begin{figure}[htpb!]%
    \centering
    \subfloat[]{{\includegraphics[]{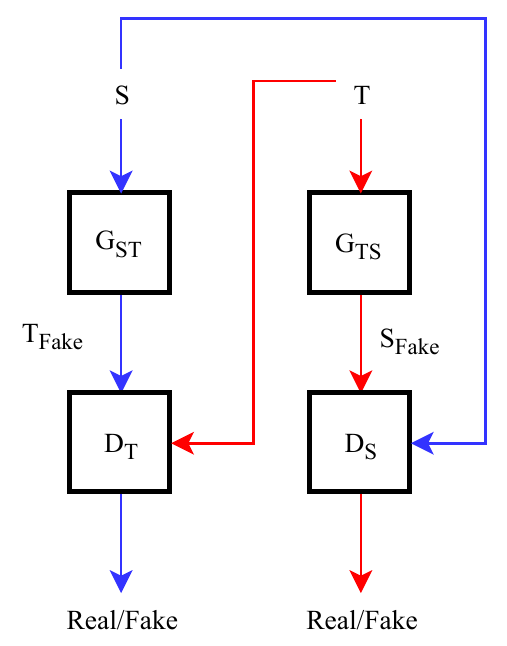} }}%
    \qquad
    \subfloat[]{{\includegraphics[]{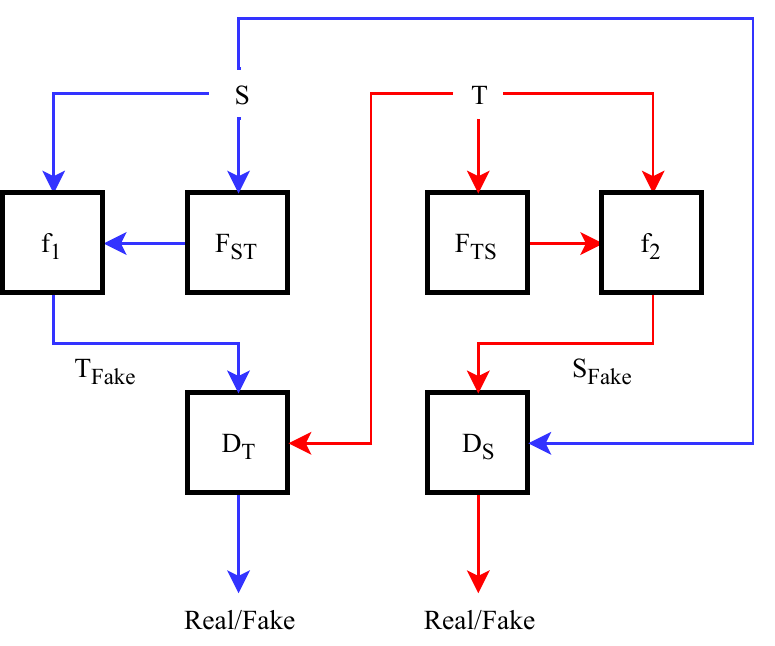} }}%
    \caption{(a): Illustration of the original Cycle-Consistent GAN (CCGAN) for image-to-image translation where the cycle consistency imposes $G_{ST}(S_{Fake})\approx T$ and $G_{TS}(T_{Fake})\approx S$. (b): The proposed Weighted Cycle-Consistent GAN (WCCGAN) inspired by Zhu et al.~\cite{zhu2017data}. Generators in our framework are $F_{ST}$ and $F_{TS}$ equivalent to $G_{ST}$ and $G_{TS}$, respectively.}
    \label{cycle_GAN_base}%
\end{figure}

The original architecture of the CCGAN~\cite{zhu2017data} is shown in Figure~\ref{cycle_GAN_base}(a) and it consists of two networks, one generator ($G$) and one discriminator ($D$) that capture data distribution and estimate the probability that a sample comes from the training data rather than $G$, respectively. $G$ in a standard GAN generates fake data from latent variables with respect to the distribution of real training data, whereas in CCGAN~\cite{zhu2017unpaired} it bijectively translates an input sample from a source $S$ to a target $T$. In other words, this type of GAN has two generators and two discriminators which are trained independently.

In this paper we focus on both paired (when $S$ is similar to $T$; or equivalently intra-class translation) and unpaired (when $S$ and $T$ are somewhat similar to each other or; equivalently inter-class translation) CCGAN. For the latter, we propose a pipeline for properly selecting source and target spectrograms with respect to the confusion matrix of the classifier. This high-level augmentation transfers structural components from the source spectrogram $S$ to the target spectrogram $T$. If the CCGAN is trained carefully, it can produce spectrogram samples that may help improve the performance of a classifier trained with such samples.

Producing realistic (natural-looking) spectrograms is not one of our priorities since any sort of spectrogram does not have much meaning for human eyes. Interestingly, spectrograms generated using our generator network may not produce samples similar to a given source, but discriminator shows reasonable sensitivity to it (matches the target label). Some examples of the generated samples are depicted in Figure~\ref{image2image}. Forcing generators to produce very similar samples will result in divergences in the cycle consistency optimization. This condition for CCGANs mostly applies for augmenting datasets to which the human eyes perceive some structure like MNIST and ImageNet datasets.

In Figure~\ref{cycle_GAN_base}(a), $G_{ST}$ and $G_{TS}$ stand for generators translating samples from $S\rightarrow T$ and $T\rightarrow S$, respectively. $D_{T}$ and $D_{S}$ denote the modules for discriminating real samples from generated fake samples from $G_{ST}$ and $G_{TS}$. This can be achieved by optimizing the following criterion:
\begin{equation}
G_{S\rightarrow T} =\arg \min_{G_{S\rightarrow T}} \max_{D_{T}} \mathcal{L}_{GAN}(G_{S\rightarrow T}, D_{T})
\end{equation}
\noindent where the loss function $\mathcal{L}_{GAN}$ is defined in Equation \ref{loss_gan}.
\begin{equation}
\begin{split}
 \mathcal{L}_{GAN}(G_{S\rightarrow T}, D_{T})= \mathop{{}\mathbb{E}_{t\sim p_{target(t)}}\left [ \log D_{T(t)}  \right ]} +\\
\mathop{{}\mathbb{E}_{s\sim p_{source(s)}}\left [ \log (1 - D_{T(t)} (G_{S\rightarrow T(s)}))  \right ]} 
\end{split}
\label{loss_gan}
\end{equation}

\noindent where $p_{target(t)}$ and $p_{source(s)}$ denote the sample distributions in the target $T$ and source $S$, respectively. The common problem with this definition of loss function is gradient vanishing which makes training and convergence almost impossible~\cite{arjovsky2017wasserstein}. To circumvent this problem, in the proposed WCCGAN architecture depicted in Figure~\ref{cycle_GAN_base}(b), we use a least-square loss function for GAN ($LSGAN$) as proposed in~\cite{mao2017least} for different domains $S$ and $T$ as given in Equations~\ref{eq:LSGAN1} and \ref{eq:LSGAN2}:

\begin{equation}
\begin{split}
\mathcal{L}_{LSGAN}(F_{S\rightarrow T}, D_{T})= \mathop{{}\mathbb{E}_{t\sim p_{target(t)}}\left [ (D_{T(t)}-1)^{2}  \right ]} +\\
\mathop{{}\mathbb{E}_{s\sim p_{source(s)}}\left [ D_{T(t)} (F_{S\rightarrow T(s)})^{2}  \right ]} 
\end{split}
\label{eq:LSGAN1}
\end{equation}

\begin{equation}
\begin{split}
\mathcal{L}_{LSGAN}(F_{T\rightarrow S}, D_{S})= \mathop{{}\mathbb{E}_{s
\sim p_{target(s)}}\left [ (D_{S(t)}-1)^{2}  \right ]} +\\
\mathop{{}\mathbb{E}_{t\sim p_{source(t)}}\left [ D_{S(s)} (F_{T\rightarrow S(t)})^{2}  \right ]} 
\end{split}
\label{eq:LSGAN2}
\end{equation}

Though these loss functions minimize the approximated Jensen-Shannon divergence between two distributions of legitimate and generated data~\cite{goodfellow2014generative}, they oversmooth the spectrograms. Oversmoothing affects the performance of the discriminator towards a wrong label other than the pre-defined target label. For rectifying this problem, we bypass the inputs to the discriminator. Hence, we add the two modules ($f_{1}$ and $f_{2}$) as depicted in Figure~\ref{cycle_GAN_base}(b), which act as weighted bypasses (identity mapping) to the discriminators. The definitions of these two modules are provided in Equations \ref{f1} and \ref{f2}.

\begin{equation}
f_{1} = c_{1}\odot S + \mu \odot F_{ST}
\label{f1}
\end{equation}

\begin{equation}
f_{2} = c_{2}\odot T + \sigma \odot F_{TS}
\label{f2}
\end{equation}

\noindent where dimensions of the generators and the input/target are bilinearly interpolated to match each other. The $\odot$ denotes the element-wise multiplication. The values of the constants $c_{1}$ and $c_{2}$, and the variables $\mu$ and $\sigma$ are obtained empirically upon several experiments. Basically, $f_{1}$ and $f_{2}$ bypass connections have two main advantages in the proposed high-level augmentation setup. First, the low-to-high compensations because the regular CCGAN (Figure~\ref{cycle_GAN_base}(a)) translates a randomly picked distribution from a low-dimension (e.g., pixel-level noisy sample) to a higher dimension which is a realistic image. Assuming that the dimension of the random drawn distribution is not very large, and no optimization overhead is involved (in the case of an optimal generator~\cite{hoang2018mgan}), then potentially the following cycle-consistency criterion can yield to a realistic fake sample:

\begin{dmath}
\mathcal{L}_{cycle}(G_{S\rightarrow T}, G_{T\rightarrow S})=\mathbb{E}_{s\sim p_{source(s)}}\left [ \left \|G_{T\rightarrow S}(G_{S\rightarrow T}(s) -s)  \right \|_{1} \right ] + \mathbb{E}_{t\sim p_{target(t)}}\left [ \left \|G_{S\rightarrow T}(G_{T\rightarrow S}(t) -t)  \right \|_{1} \right ] 
\label{lcyc1}
\end{dmath}

\noindent where $\left \| . \right \|_1$ is the $L_{1}$ norm. This might converge to a saddle point (where the minimax game in GAN is over) when the Kullback-Leibler divergence $\mathbb{KL}(p_{source(s)}, p_{target(t)}) \approx \mathbb{KL}(p_{target}, p_{source})$. In other words, the similarity between the source and the target distribution should be high. When the similarity between samples is not high enough especially when they have been drawn from different classes, Equation \ref{lcyc1} can no longer result in realistic fake images. $f_{i}$ bypasses can overcome this problem by providing more information from a given legitimate input.

The second advantage of embedding $f_{1}$ and $f_{2}$ into the proposed WCCGAN is the ability of sharpening features that may have been oversmoothed during translation (especially in the discriminator domain). Finally, the total loss criterion which is optimized in our augmentation scenario is given in Equation~\ref{myLosst}:

\begin{dmath}
\mathcal{L}_{total}({F_{S\rightarrow T}}, {F_{T\rightarrow S}}, D_{S}, D_{T})=\mathcal{L}_{LSGAN}(F_{S\rightarrow T}, D_{T}) + \mathcal{L}_{LSGAN}(F_{T\rightarrow S}, D_{S}) + \alpha \mathcal{L}_{cycle}(F_{S\rightarrow T}, F_{T\rightarrow S}) 
\label{myLosst}
\end{dmath}

\noindent where $\alpha$ is a scaling parameter for balancing the cycle whose value is also set manually upon experiments.

\subsection{ConvNet Architecture for the Weighted Cycle-Consistent GAN}
\begin{figure*}[htpb!]
  \centering
	\includegraphics[width=\textwidth]{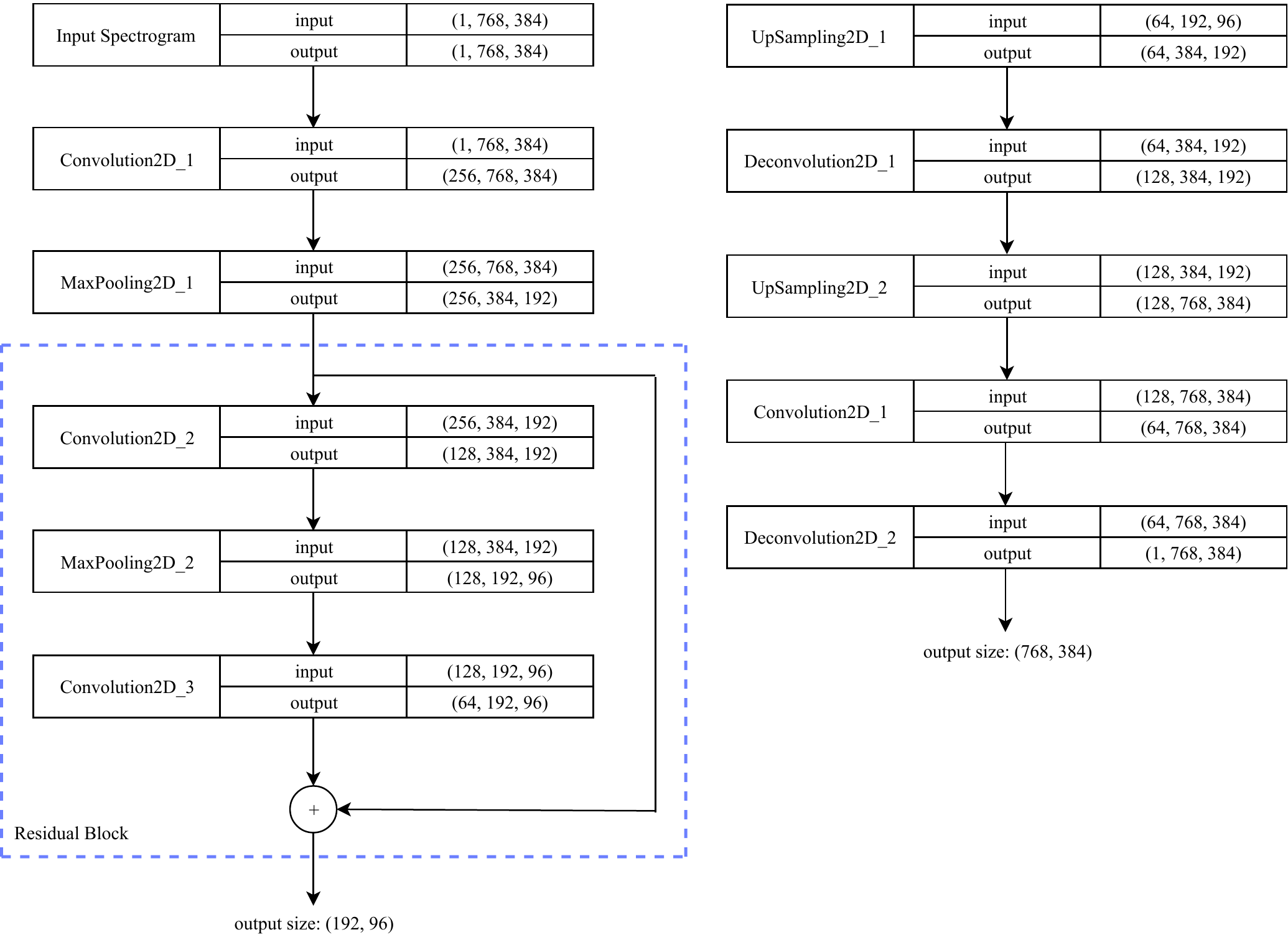}
  \caption{Generator architectures for DWT spectrograms: \textbf{left:} $F_{S\rightarrow T}$, and \textbf{right:} $F_{T\rightarrow S}$. Values inside of parentheses indicate the number of filters, height, and width of the spectrogram, respectively.}
  \label{FST}
\end{figure*}

Assuming that we generate DWT spectrograms of 768$\times$384 pixels, for high-level augmentation using the WCCGAN, we propose the architectures illustrated in Figure~\ref{FST} for the generators ($F_{S\rightarrow T}$, $F_{T\rightarrow S}$). We started with a complex ConvNet model based on the AlexNet architecture for all four networks (generators and discriminators) of Figure~\ref{cycle_GAN_base}(b), and we simplified this network by removing some layers which resulted in a simpler ConvNet architecture with 30\% fewer parameters than AlexNet. Furthermore, when the architectures of discriminators and generators are similar, the cycle-consistency loss function follows a smooth and convex descending track. Therefore, we proposed two equivalent discriminators for both source-to-target and target-to-source mappings. The residual network shown in Figure~\ref{FST} may have from three to seven residual blocks~\cite{he2016deep}, depending on the dataset. Each residual block contains two convolution layers and one bypassing residual connection. In all the layers depicted in Figure~\ref{FST} the convolution layers have receptive field of 3$\times$3 and stride 1$\times$1. Also, the sizes of the generated outputs in the residual blocks are bilinearly interpolated to match each other. These architectures are not general and they need to be adapted depending on the type of problem and dataset.

\begin{figure}[htpb!]
  \centering
  \includegraphics[width=.45\textwidth]{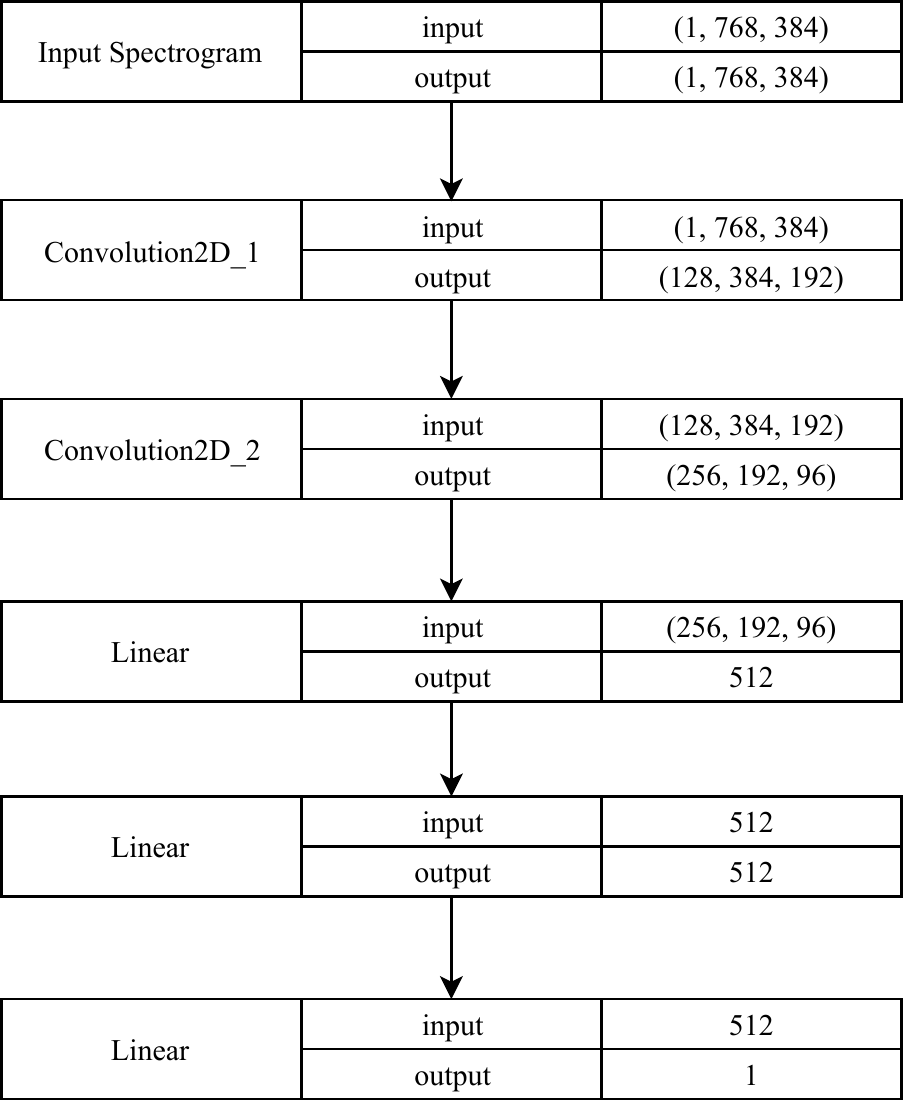}
  \caption{Network architecture for $D_{T}$ and $D_{S}$.}
  \label{DT}
\end{figure}

For discriminator functions $D_{T}$ and $D_{S}$ we use a single architecture as depicted in Figure~\ref{DT}. In the proposed architecture we also have receptive field of 3$\times$3 and strides 1$\times$1 and 2$\times$2 for the first and second convolution layers, respectively. There is no generic way for determining an optimal structure for these two networks and we have basically rely on our initial experiments on UrbanSound8k dataset. Changing the structures of these networks might affect the performance of image-to-image translation and it probably needs additional modifications/tuning of the hyperparameters. Therefore, we used the same architecture for the other datasets, but we have optimized the hyperparameters. Even if such an architecture is not customized to the other datasets, we have achieved good results as we show in Section \ref{sec:exp}.

\section{Unsupervised Feature Learning and Classification}
\label{sec:featlearn}
The proposed approach for classifying DWT spectrograms is based on an unsupervised feature learning approach. The motivation behind proposing a shallow approach instead of a deep architecture as a front-end classifier is twofold. First, it has been shown that advanced deep neural networks such as AlexNet, GoogLeNet and other recent architectures are highly vulnerable to adversarial attacks as they can predict wrong labels with high confidence~\cite{esmaeilpour2019robust, szegedy2013intriguing}. Secondly, conventional classifiers such as SVMs and RFs, which learn from handcrafted features are considerably more robust against such adversarial attacks than deep learning models~\cite{esmaeilpour2019robust}. Taking advantage of these two facts, we propose a conventional data-driven model as front-end classifier and use a generative model based on a deep architecture as a back-end classifier for data augmentation purposes only. Therefore, the deep architecture helps the front-end classifier to learn more discriminant boundaries. In this section, we present how to extract features from spectrograms and learn a codebook of representative codewords.

\subsection{Feature Encoding}
For extracting feature from the spectrograms, the speeded up robust feature (SURF ~\cite{bay2006surf}) is implemented, which is the modified version of the scale invariant feature transform (SIFT)~\cite{lowe1999object} by fast approximation of Hessian matrix (for encoding principal curvatures at each point of interest) and producing integral images from spectrograms. Upon several experiments, SURF visual words from DWT provide us with better feature vectors compared to MFCC visual words which have been studied for music and environmental sound classification~\cite{salamon2015unsupervised,vaizman2014codebook}. In Section \ref{sec:exp} we provide some additional result in extracting SURF features from MFCC.  

Each integral image represents the summation of the spectrogram pixels of a rectangular region with different sizes to produce local features. Using the box filter (for Gaussian approximation), SURF approximates the location and scale of each point of interest by using the determinant of the weighted Hessian matrix as the following.

\begin{equation}
H(\texttt{p},\sigma) \approx \begin{bmatrix}
\hat{L}_{xx}(p,\sigma) & \hat{L}_{xy}(p,\sigma) \\ 
\hat{L}_{xy}(p,\sigma) & \hat{L}_{yy}(p,\sigma)
\end{bmatrix}
\end{equation}
\begin{equation}
\det(H(\texttt{p},\sigma)) = \hat{L}_{xx}(p,\sigma)\hat{L}_{yy}(p,\sigma)-[0.9(\hat{L}_{xy}(p,\sigma))]^{2}
\end{equation}

\noindent where $\hat{L}_{. .}(p,\sigma)$ is the convolution of the second derivative of Gaussian with the spectrogram $S(x,y)$ at point $x$, and $\sigma$ is the Gaussian scale (scale at which the point has been detected). After locating the interest points in space and scale, the SURF descriptor can be generated.

Assuming once again that we generate DWT spectrograms of 768$\times$384 pixels, we divide each spectrogram into 16 sub-regions (4$\times$4 grids of size 4$\times$4) and compute Haar wavelet responses for obtaining orientation of interest points. In each sub-region, we compute a four-element descriptor vector as given by Equation~\ref{descriptor}:

\begin{equation} 
\texttt{descriptor}_{subregion}=
\begin{bmatrix}
\sum dx, \sum dy, \sum\left | dx \right |, \sum\left | dy \right |
\end{bmatrix}
\label{descriptor}
\end{equation}

The length of the regional feature descriptor is 16$\times$4 which is represented by a 64-dimensional vector. These values are determined empirically on UrbanSound8k dataset and they do not change during the implementation or across datasets. The majority of the settings for feature extraction are the default parameters of the OpenCV Library. For detecting interest points, a blob detector based on the Hessian matrix is implemented. Different Hessian threshold values have been evaluated, ranging from 250 to 1~000 on 15\% of randomly selected samples of the dataset with four trials and 400 was set as a fair average threshold with respect to the performance of our classifier. Roughly, about 900 keypoints have been detected in each spectrogram. We have employed non-maximum suppression strategy with threshold of 0.6 to rectify the problem of detecting too many features. We skipped subregions in which SURF could not detect any feature. More details are presented in Section \ref{sec:exp}.

High resolution spectrograms to some extent can help SURF to extract more meaningful features but does not necessarily increase the performance in classification. Our main emphasis in this paper is the high-level augmentation which basically maps one sample to another aiming at increasing intra-class similarity and inter-class dissimilarity, regardless of the quality of the spectrogram. Resizing resolution of spectrograms which perhaps change the size of sub-regions, slightly affects the quality of the extracted features. For spectrograms of higher resolution (for instance 1152$\times$576), we suggest increasing the dimension of feature vectors to 128 as our initial experiments have shown its positive impact on the final classification performance. In the next step, we learn a codebook of representative codewords (a.k.a visual words) from such feature vectors.

\subsection{Organizing Visual Words into a Codebook Using Spherical {\it K}-Means++}
The number of feature vectors extracted from the spectrograms is tremendously high and this negatively affects classifier`s performance. Therefore, representing these vectors with respect to their similarities into centers and organizing them into a codebook can considerably improve the classification process.

We use the $K$-Means++ algorithm~\cite{arthur2007k} as an unsupervised feature learning for organizing codewords. This clustering algorithm is adapted from the traditional $K$-Means algorithm where $K$ denotes the number of potential seeds (centroids). This value is usually larger than the dimensionality of the audio data. The main advantage of the $K$-Means++ algorithm over the traditional $K$-Means algorithm is that it uses a weighted probability distribution over the data point (feature vector in our case) sub-manifold(s) with probability proportional to its squared distance to its neighbors. This is very useful in our case since feature vectors are not extracted from solid images. Similar to the traditional $K$-Means, the $K$-Means++ algorithm has a super polynomial structure and it might result in null seeds for similar data points~\cite{dhillon2001concept}. One possible solution provided for $K$-Means is adding an extra optimization constraint by binding seeds to have a unit $L_{2}$ norm which forces the centroid to roll over a unique sphere. This algorithm is called spherical $K$-Means~\cite{coates2012learning}. By taking advantage of this extra constraint and embedding it into the $K$-Means++ clustering algorithm, spherical $K$-Means++~\cite{endo2015spherical} turns out. The performance of standard spherical $K$-Means is studied for specific forms of environmental sound dataset with quite small cardinality~\cite{stowell2014automatic}. It has been proven that, this clustering algorithm produces competitive results with cutting-edge clustering and other advanced supervised classifiers~\cite{dieleman2013multiscale}. Adding a spherical constraint in the distance objective function of $K$-Means usually results in improving the consistency in producing centroids. 

Considering the feature vectors of an input spectrogram represented as a $X_{m,n}$ matrix where $m$ and $n$ denote the number of feature vectors and their dimensionality in the form of 1$\times$n, respectively ($1 \le i \le m$ and $1 \le j \le n$). In Equation~\ref{eq:zi} we define $z_{i}$ for storing the assigned value (mean of centroids) of our $K$ clusters which forms the matrix $Z$. Finally, our codebook is defined as $V \in \mathbb{R}^{n\times K}$.

\begin{equation}
z_{j}^{i} := \left\{\begin{matrix}
V^{j} x^{i^\top} & \mathrm{if} \, j= \arg \underset{l}{\textrm{max}} \begin{vmatrix}
V^{l} x^{i^\top} \end{vmatrix}_{j,i} \textrm{and} \quad  p(x)\sim cp(d^{2}(x^{i}))\\ 
0 &  \mathrm{otherwise}
\end{matrix}\right.
\label{eq:zi}
\end{equation}

\noindent where $x^i$ is a row from $X$ and $c$ is a constant value for weighting the square distance of each $x^i$ to its nearest center. Specifically, $d$ and $p$ denote the distance between two feature vectors and their joint probability distribution, respectively, and $\top$ indicates matrix transposition. More details about the basics of spherical $K$-Means algorithm is provided in~\cite{coates2012learning}. Finally, the two operations of Equation~\ref{eq:upcen} update the centroids and normalize them by $L_2$ norm, respectively. The centroids can be randomly normalized following a normal distribution. The codebook matrix $V$ contains $K$ organized clusters that we use to encode the training data and train a classifier. 

\begin{equation}
V := XZ^\top+V, \quad \quad V^{j}:=\frac{V^{j}}{\left \| V^{j} \right \|_{2}} \quad \forall j
\label{eq:upcen}
\end{equation}

\subsection{Classification}
For classifying the code vectors encoded against the codebook, we have considered the most performing conventional approaches, namely SVM with different kernels (linear, polynomial, radial basis functions) and random forest (RF). We decided to use a RF as our front-end classifier based on the recognition accuracy. We use the random forest (RF) algorithm~\cite{breiman2001random} with a different number of trees. This algorithm is an estimator which fits some decision trees on different sub-samples of given code vectors via averaging. We train this algorithm with different sizes of trees (estimators) with respect to the dimensions of the generated code vectors. For splitting a random tree node, the Gini impurity criterion is used as follows:

\begin{equation}
G=\sum_{i=1}^{n}p_{i}(1-p_{i})
\end{equation}

\noindent where $n$ denotes the number of classes in the target variable and $p_{i}$ is the ratio of picking a random sample from class $i$. The maximum depth of trees varies from 16 to 64 with respect to the type of codebook. Specifically, for code vectors associated with long audio recordings, we use deeper trees. 
The minimum number of samples required to split an internal node is set to 0.02$\times m$ where $m$ stands for the number of samples per class. This classifier has shown great potential for classifying code vectors~\cite{salamon2014dataset}. Upon our initial experiments, we have noticed that spherical biding of code vectors for the $K$-Means++ outperforms the standard one.

\section{Experimental Results}
\label{sec:exp}
We assess the performance of the proposed approach in four environmental sound datasets: UrbanSound8k, ESC-10, ESC-50, and DCASE-2017. The first dataset includes 8~732 audio samples of up to four seconds in duration distributed in 10 classes: air conditioner (AI), car horn (CA), children playing (CH), dog bark (DO), drilling (DR), engine idling (EN), gunshot (GU), jackhammer (JA), siren (SI), and street music (SM). The ESC-50 includes 2~000 samples of 5-second duration distributed in 50 classes including major groups of animals, natural sound capes and water sounds, human non-speech sounds, domestic sounds, and exterior noises. The ESC-10 is a subset of ESC-50 which includes 400 excerpts arranged in 10 classes: dog bark, rain, sea waves, baby cry, clock tick, person sneeze, helicopter, chainsaw, rooster, fire crackling. Finally, DCASE-2017 consists of 4~680 10-second audio samples from 15 classes: bus, cafe, car, city center, forest path, grocery store, home, lakeside beach, library, metro station, office (multiple persons), residential area, train, tram, and urban park. Though the cardinality of samples per class in UrbanSound8k is not balanced, such a dataset contains the most challenging environmental sounds in real life compared to the other three datasets in terms of including different sound production mechanisms. 

For low-level data augmentation, there is no automatic approach for tuning the pitch-shifting hyperparameter ($t$) and this depends on the type of audio signal, as mentioned in Section 2.1. Therefore, we have carried out a grid search starting by $t \in \{0.6, 0.75, 0.9, 1.1, 1.25, 1.4\}$ as suggested in~\cite{boddapati2017classifying} and we have found that more than 25\% signal compression ($t < 0.75$) does not increase the F1 score of our approach. This makes sense because pitch-shifting with $t<1$ is a lossy operation and it might increase the chance of losing pivotal frequency components. Overall, for pitch-shifting with $t <1$ we kept only the two most influential values (0.75 and 0.9). For pitch-shifting with $t >1$ we started with 1.1 and gradually increase it by 0.05 displacement to the margin of 65\% signal stretching compared to the original signal. Stretching signals with $t >1.65$ did not result in a positive effect on the performance of the front-end classifier. We speeded up all audio samples with $t \in \{1.1, 1.15, 1.2,1.25, 1.3, 1.35, 1.4, 1.45, 1.5, 1.55, 1.6, 1.65\}$ and ranked them with respect to the F1 score measured for the front-end classifier. We finally kept $t \in \{1.15, 1.5\}$ for stretching the audio signal as the rest did not show any considerable improvement.
Therefore, using static pitch shifting scales of 0.75, 0.9, 1.15, and 1.5, we ended up with an augmented dataset of 43~660, 10~000, 2~000 and 23~400 samples for UrbanSound8k, ESC-50, ESC-10 and DCASE-2017 datasets, respectively.

For each audio sample in the augmented datasets, we generate the DWT spectrograms by setting the sampling frequency of 8 kHz for ESC-10 and UrbanSound8k datasets, and 16 kHz to ESC-50 and DCASE-2017 datasets. Besides, we also set the frame length to 50 ms for ESC-10 and UrbanSound8k, 30 ms for ESC-50, and 40 ms for DCASE-2017 with a fixed overlapping size of 50\%~\cite{boddapati2017classifying}. Therefore, each audio samples is now represented by a DWT spectrogram of 768$\times$384 pixels. Empirically, this resolution provides a fair tradeoff between information content (in terms of feature vectors) and dimensionality. Each spectrogram undergoes to the enhancement step and next we apply the high-level data augmentation using the proposed WCCGAN.

The proposed WCCGAN employs the ConvNets presented in Figure~\ref{FST}, which have normalized convolution layers by applying instance normalization~\cite{ulyanovinstance} technique followed by leaky ReLU activation function with slope $0.3$. We used Glorot weight normalization algorithm for improving learning. For discriminator functions $D_{T}$ and $D_{S}$ we use a single architecture as depicted in Figure~\ref{DT}. In the proposed architecture we set the receptive field to 3$\times$3 and strides are set to 1$\times$1 and 2$\times$2 for the first and second convolution layers, respectively. In this case, we used ReLU activation function and batch normalization~\cite{gulrajani2017improved}. These four networks are trained in four parallel GPUs GTX580 based on an implementation proposed in~\cite{krizhevsky2012imagenet}. We applied early stopping policy for training these networks and the total number of epochs for training each network is shown in Table~\ref{epochs}.

\begin{table}[htpb!]
\footnotesize
\renewcommand{\arraystretch}{1.3}
\centering
\caption{The total number of training epoch for four networks generators and discriminators.}
\label{epochs}
\begin{tabular}{|c|c|c|c|c|}
\hline
 & \multicolumn{4}{c|}{\# of Training Epochs}\\ \cline{2-5}

Dataset & $F_{S\rightarrow T}$ & $F_{T\rightarrow S}$ & $D_{S}$ & $D_{T}$ \\ \hline
ESC-10       & 123  & 107  & 104  &  91 \\ \hline
ESC-50       &  136 & 118  & 109  &  116 \\ \hline
UrbanSound8k & 112  & 106  & 97  &  45 \\ \hline
DCASE-2017   &  213 & 143  & 90  &  102 \\ \hline
\end{tabular}
\end{table}

The tentative values for $c_{1}$, $c_{2}$, $\mu$, $\sigma$, and $\alpha$ in Equations~\ref{f1}, \ref{f2}, and \ref{myLosst} for each dataset are shown in Table~\ref{Detail_apram}. There is no deterministic approach to adjust such hyperparameters of the WCCGAN. Moreover, there is no guarantee that such hyperparameters are properly set, as they result from exploratory experiments where we empirically modified them up to see a good track in sample generation and detection. In all experiments the main criterion was achieving the best epoch before overtraining generators and discriminators using early stopping. We have changed the hyperparameters almost randomly to get the best epoch. Since $F_{S\rightarrow T}$ is stronger than $F_{T\rightarrow S}$ due to the residual blocks, we intentionally increase the weight of the latter generator for all the datasets. This is the main reason for having higher values for $\sigma$ compared to $\mu$ in all experiments. Hyperparameters $c_{1}$ and $c_{2}$ are weights for source and target samples respectively. Hence, except for the DCASE-2107 dataset, we tried to keep the summation of these weights close to one to ensure balanceness. The hyperparameter $\alpha$ keeps the cycle consistency and we noticed that it should not exceed 0.45 for the proposed setup as higher values do not lead to convergence of the generators. Table~\ref{Detail_apram} shows the hyperparameter values found by a basic and non-optimal local random search that attempts to find the models that produce the best F1 score in terms of a minimum number of epochs. Once the best hyperparameters have been found, we applied perturbations of $\pm$2\%, $\pm$5\%, and $\pm$10\% to assess the sensitivity of the WCCGAN in respect to these hyperparameters. The F1 score of the discriminator networks has been computed for each perturbation applied on the hyperparameters of Table \ref{Detail_apram} which resulted in a noticeable performance drop, ranging from 2.4\% to 12.6\%, depending on the dataset and hyperparameter. As expected, these hyperparameters have a great influence in the performance of the WCCGAN because they are tuned upon a local search to allow the WCCGAN to produce spectrograms with low inter-class and high intra-class similarity. Among all these hyperparameters, $\alpha$ is the most sensitive one as it controls the consistency. In other words, this hyperparameter leverages the cycle-consistency loss between generators and acts to some extent as a regularizer for the generators.

\begin{table}[htpb!]
\footnotesize
\centering
\renewcommand{\arraystretch}{1.3}
\caption{Values for the hyperparameters of Equations~\ref{f1}, \ref{f2}, and \ref{myLosst}, obtained by a local random search.}
\label{Detail_apram}
\begin{tabular}{|c|c|c|c|c|c|}
\hline
& \multicolumn{5}{c|}{Hyperparameter Values}\\\cline{2-6}
Dataset         & $c_{1}$ & $c_{2}$ & $\mu$ & $\sigma$ & $\alpha$ \\ \hline
ESC-10       	&  0.49 &  0.67 & 0.02  & 0.76  & 0.23\\ \hline
ESC-50       	&  0.39 &  0.68 & 0.12  & 0.58  & 0.19 \\ \hline
UrbanSound8k 	&  0.62  & 0.36  &  0.14 & 0.57  & 0.03\\ \hline
DCASE-2017      &  0.03 &  0.21 & 0.18  & 0.43  & 0.31 \\ \hline
\end{tabular}
\end{table}

\begin{figure*}[htpb!]
  \centering
  \includegraphics[width=0.8\textwidth]{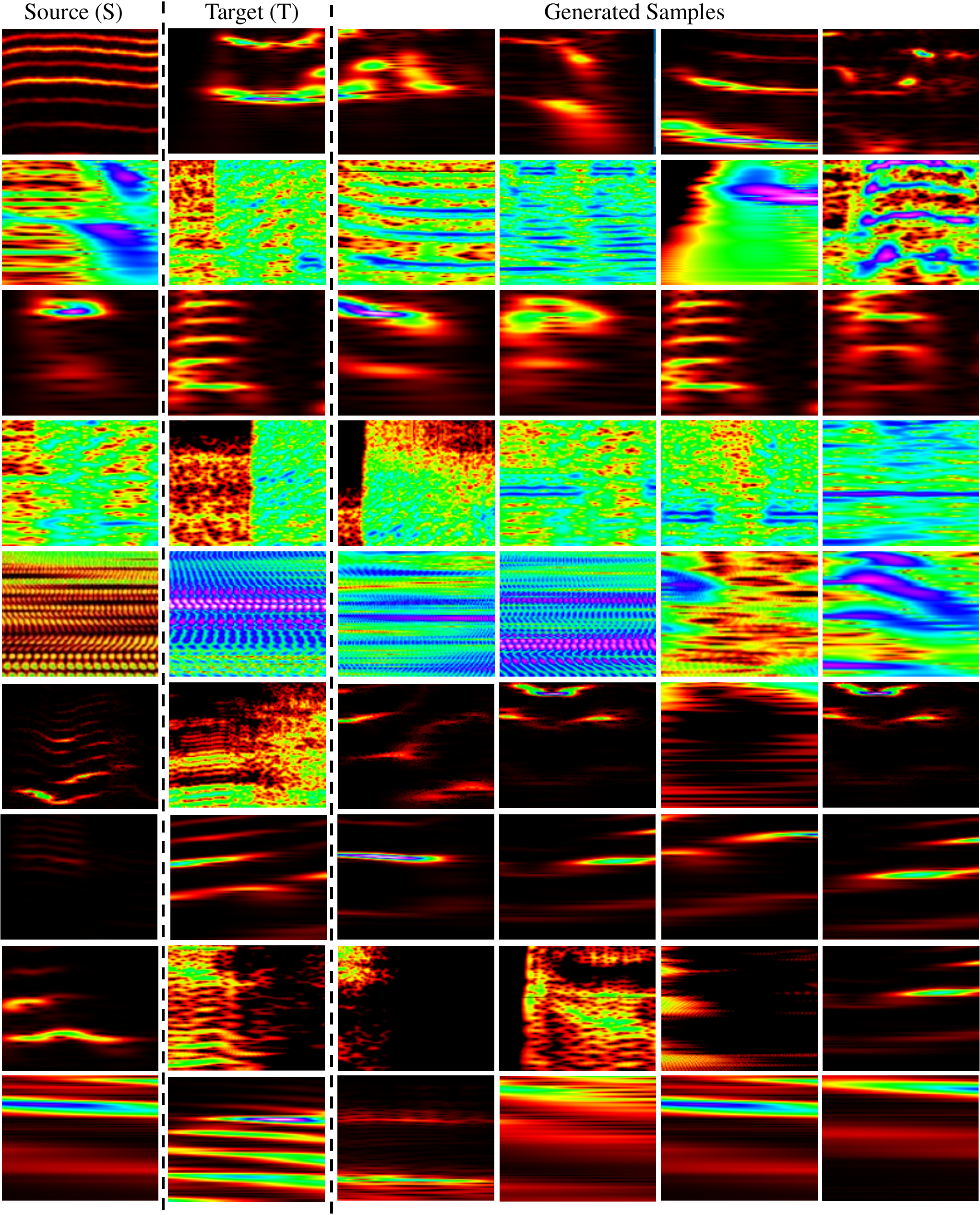}
  \caption{Generated spectrograms using the WCCGAN for randomly drawn sources ($S$) and targets ($T$). The $S$s and $T$s shown in the top four rows indicate intra-class image-to-image translation. Specifically, UrbanSound8k ($S=T$: sea waves), ESC-10 ($S=T$: person sneeze), ESC-50 ($S=T$: pouring water), and DCASE-2017 ($S=T$: office). Sources and targets for inter-class translation are shown in the five bottom rows as in UrbanSound8k ($S$: sea waves, $T$: rain), ESC-10 ($S$: person sneeze, $T$: helicopter), ESC-50 ($S$: wind, $T$: pouring water), and DCASE-2017 ($S$: cafe, $T$: office).}
  \label{image2image}
\end{figure*}

In order to produce more structural spectrograms from source $S$ to target $T$ and make the loss functions converge, we need to have an idea of the inter-class relation between samples. For such an aim, we randomly pick samples to train a RF algorithm on spectrograms without high-level data augmentation featuring different number of trees from 500 to 3~000. Table~\ref{conf_without_gan_updated} shows the confusion matrices for the RF trained with the UrbanSound8k dataset without high-level data augmentation. The values in Table~\ref{conf_without_gan_updated} can also be interpreted as similarity among classes. For instance, class "EN" has high similarity with class "AI" because the classifier has been misclassified samples from the class "AI" as class "EN" in 14\% of the cases. Therefore, we set the source and target classes in Figure~\ref{cycle_GAN_base} to $S$="AI" and $T$="EN", respectively. We use the same procedure for all classes. In addition to intra-class image-to-image translation, we augment the DWT spectrograms in inter-class manner as well. We randomly select 50\% of samples within a class as the source and the remaining 50\% as the target classes. Overall, we increase the size of the datasets with extra 1~500, 2~000, 5~000 and 4~500 samples for ESC-10, ESC-50, UrbanSound8k and DCASE-2017, respectively. Some visual examples of the generated spectrograms using the WCCGAN are shown in Figure~\ref{image2image}. This figure shows the high capability of the WCCGAN for producing structurally similar spectrograms even when the source and target are not similar to each to the human eye perspective.

\begin{table}[ht]
\centering
\footnotesize
\caption{Confusion matrix of the proposed classification approach without high-level augmentation on the UrbanSound8k dataset. Values in bold indicate the best recognition accuracy in a 5-fold cross validation setup.}
\begin{tabular}{|c|c|c|c|c|c|c|c|c|c|c|}
\hline
   & AI     & CA     & CH     & DO     & DR     & EN     & GU     & JA     & SI              & SM     \\ \hline
AI & $\textbf{0.68}$ & $0.00$ & $0.02$ & $0.01$ & $0.05$ & $0.14$ & $0.01$ & $0.04$ & $0.02$ & $0.03$ \\ \hline
CA & $0.00$ & $\textbf{0.77}$ & $0.02$ & $0.02$ & $0.00$ & $0.00$ & $0.00$ & $0.05$ & $0.07$          & $0.07$ \\ \hline
CH & $0.07$ & $0.05$ & $\textbf{0.31}$ & $0.09$ & $0.04$ & $0.03$ & $0.02$ & $0.04$ & $0.15$          & $0.20$ \\ \hline
DO & $0.06$ & $0.04$ & $0.03$ & $\textbf{0.68}$ & $0.04$ & $0.02$ & $0.03$ & $0.00$ & $0.05$          & $0.05$ \\ \hline
DR & $0.02$ & $0.04$ & $0.02$ & $0.02$ & $\textbf{0.74}$ & $0.01$ & $0.01$ & $0.10$ & $0.04$          & $0.00$ \\ \hline
EN & $0.04$ & $0.00$ & $0.03$ & $0.02$ & $0.01$ & $\textbf{0.78}$ & $0.02$ & $0.06$ & $0.01$          & $0.03$ \\ \hline
GU & $0.00$ & $0.02$ & $0.00$ & $0.03$ & $0.00$ & $0.00$ & $\textbf{0.95}$ & $0.00$ & $0.00$          & $0.00$ \\ \hline
JA & $0.01$ & $0.01$ & $0.00$ & $0.00$ & $0.05$ & $0.03$ & $0.00$ & $\textbf{0.90}$ & $0.00$          & $0.00$ \\ \hline
SI & $0.03$ & $0.06$ & $0.03$ & $0.02$ & $0.02$ & $0.01$ & $0.03$ & $0.01$ & $\textbf{0.78}$          & $0.01$ \\ \hline
SM & $0.03$ & $0.08$ & $0.06$ & $0.09$ & $0.08$ & $0.08$ & $0.01$ & $0.04$ & $0.06$          & $\textbf{0.47}$ \\ \hline
\end{tabular}
\label{conf_without_gan_updated}
\end{table}

\begin{table}[ht]
\centering
\footnotesize
\caption{Confusion matrix of the proposed classification approach with WCCGAN augmentation on the UrbanSound8k dataset. Values in bold indicate the best recognition accuracy in a 5-fold cross validation setup.}
\begin{tabular}{|c|c|c|c|c|c|c|c|c|c|c|}
\hline
   & AI     & CA     & CH     & DO     & DR     & EN     & GU     & JA     & SI              & SM     \\ \hline
AI & $\textbf{0.89}$ & $0.01$ & $0.00$ & $0.02$ & $0.01$ & $0.04$ & $0.00$ & $0.01$ & $0.02$ & $0.00$ \\ \hline
CA & $0.01$ & $\textbf{0.92}$ & $0.00$ & $0.01$ & $0.02$ & $0.00$ & $0.01$ & $0.01$ & $0.00$          & $0.02$ \\ \hline
CH & $0.00$ & $0.01$ & $\textbf{0.91}$ & $0.03$ & $0.00$ & $0.01$ & $0.00$ & $0.00$ & $0.01$           & $0.03$ \\ \hline
DO & $0.00$ & $0.00$ & $0.01$ & $\textbf{0.96}$ & $0.01$  & $0.00$ & $0.00$ & $0.00$ & $0.01$          & $0.01$ \\ \hline
DR & $0.00$ & $0.01$ & $0.00$ & $0.02$ & $\textbf{0.95}$ & $0.01$ & $0.00$ & $0.01$ & $0.00$          & $0.00$ \\ \hline
EN & $0.01$ & $0.00$ & $0.00$ & $0.01$ & $0.00$ & $\textbf{0.96}$ & $0.01$ & $0.00$ & $0.00$          & $0.01$ \\ \hline
GU & $0.01$ & $0.00$ & $0.00$ & $0.00$ & $0.01$ & $0.00$ & $\textbf{0.97}$ & $0.01$ & $0.00$          & $0.00$ \\ \hline
JA & $0.02$ & $0.00$ & $0.00$ & $0.00$ & $0.00$ & $0.00$ & $0.00$ & $\textbf{0.98}$ & $0.00$          & $0.00$ \\ \hline
SI & $0.00$ & $0.01$ & $0.03$ & $0.00$ & $0.01$ & $0.00$ & $0.00$ & $0.00$ & $\textbf{0.95}$          & $0.00$ \\ \hline
SM & $0.00$ & $0.01$ & $0.02$ & $0.01$ & $0.00$ & $0.00$ & $0.01$ & $0.00$ & $0.00$          & $\textbf{0.95}$ \\ \hline
\end{tabular}
\label{conf_with_gan_updated}
\end{table}

After finishing both inter- and intra-class data augmentation processes, we train again the RFs on the augmented dataset, considering different number of trees. The best number of trees for ESC-10, ESC-50, UrbanSound8k, and DCASE-2017 were obtained at 2~000, 1~864, 2~500, and 2~496, respectively with minimum AUC metrics (one-vs-all). Table~\ref{conf_with_gan_updated} shows the performance of the learned trees on the UrbanSound8k dataset augmented with the proposed WCCGAN. The results are highly improved compared to the trees trained on codebooks learned without high-level data augmentation (Tables~\ref{conf_without_gan_updated}). This shows the importance of high-level data augmentation for extracting more discriminating features.

\begin{figure*}[htpb!]
  \centering
  \includegraphics[width=1\textwidth]{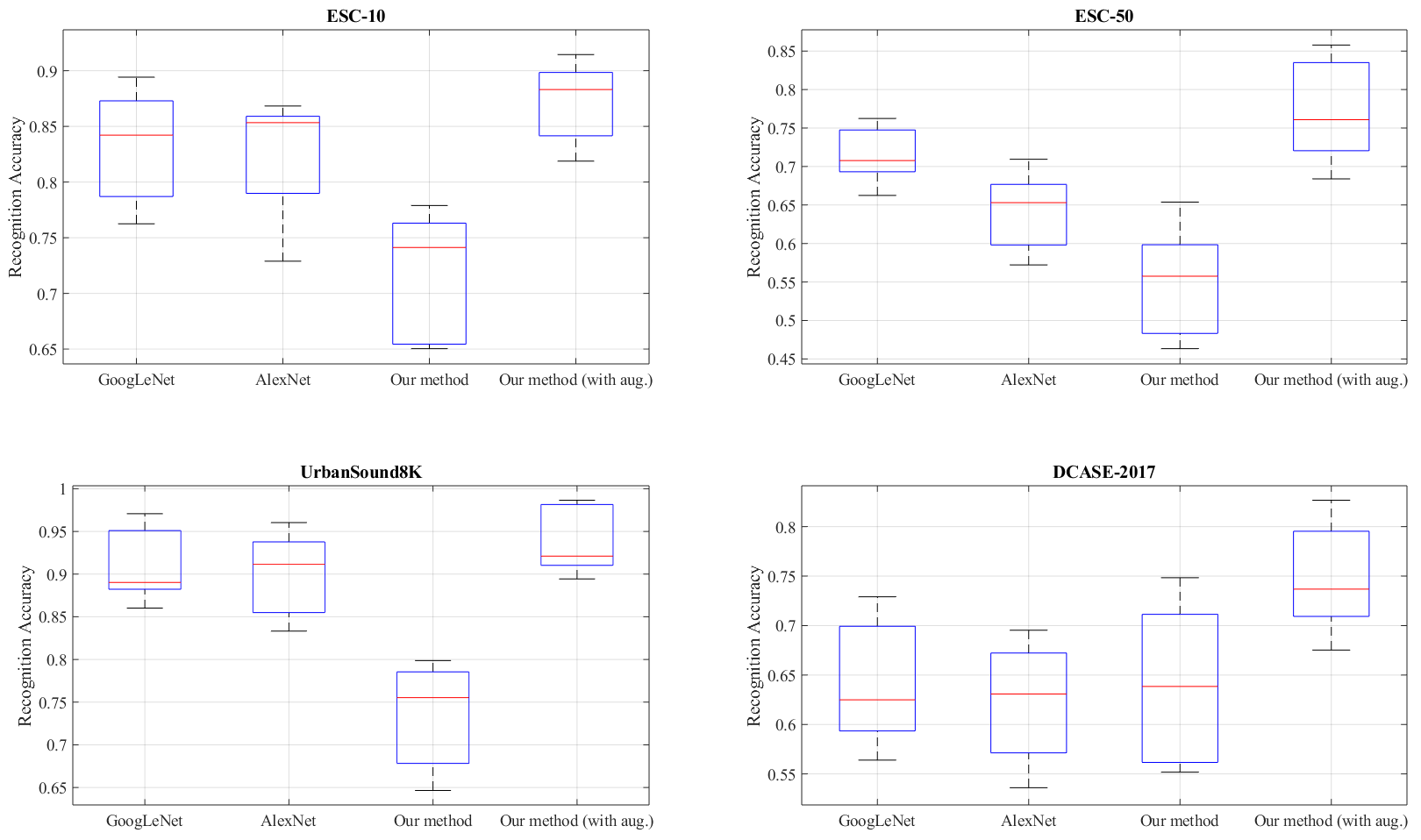}
  \caption{Box-plots of the approaches from Table~\ref{Rec_acc} in a 5-fold cross validation setup for ESC-10, ESC-50, UrbanSound8k and DCASE-2017 datasets.}
  \label{boxplot1}
 \end{figure*}

Table~\ref{Rec_acc} compares the performance of the proposed classification approach to the state-of-the-art pre-trained classifiers (AlexNet and GoogLeNet) on environmental sound datasets following the transfer learning and fine-tuning strategies explained in~\cite{kumar2018knowledge}. It is worth mentioning that, these two pre-trained networks have been fine-tuned on the 2D aggregation (pooling) of STFT, MFCC, and CRP. As Table~\ref{Rec_acc} shows, our approach outperforms both deep learning models on all environmental sound datasets. One clear outcome of Table~\ref{Rec_acc} is that the GAN theory could help us not only to build robust classifiers, but also to highlight another traditional classifier's performance. Furthermore, for a better comparison of the performances, the box-plots of these classifiers are shown in Figure~\ref{boxplot1}. With respect to these box-plots for all the four benchmarking datasets, the proposed approach using the WCCGAN architecture for high-level data augmentation achieved the highest maximum, mean, minimum, and median accuracy. These plots also confirm that the proposed approach together with WCCGAN outperforms AlexNet and GoogLeNet since it provides the highest statistical measures except for the ESC-50 dataset; and there are no outliers. In order to investigate the statistical significance of the recognition performances reported in Table~\ref{Rec_acc}, we used Friedman’s test which is the non-parametric version of the one-way ANOVA with some limited repeated measures~\cite{hogg1987engineering}. Upon 19 runs (degrees of freedom), we could reach the $p$-value of 0.05 on average, which shows the high performance of the proposed approach.

\begin{table*}[htpb!]
\footnotesize
\renewcommand{\arraystretch}{1.3}
\centering
\caption{Comparing the mean accuracy of the proposed approach with and without high-level augmentation (DA) with GoogLeNet and AlexNet. Comparison has been made in a 5-fold cross validation setup. The best results are shown in bold faces.}
\label{Rec_acc}
\begin{tabular}{|c|c|c|c|c|}
\hline
& \multicolumn{4}{c|}{Mean Accuracy}\\ \cline{2-5}
Dataset            &  &  & \multicolumn{2}{c|}{Proposed Approach } \\ \cline{4-5}
& GoogLeNet & AlexNet & Without DA & With DA \\ \hline
ESC-10 & 0.83 & 0.83 & 0.72 &  \textbf{0.87}   \\ \hline
ESC-50 & 0.71   & 0.64    &  0.55 &    \textbf{0.77} \\ \hline
UrbanSound8k &  0.91   &  0.90   &  0.73  &    \textbf{0.94} \\ \hline
DCASE-2017 &  0.64   &  0.62   &  0.66 &    \textbf{0.76}  \\ \hline
\end{tabular}
\end{table*}

Even if the current state-of-the-art is based on pre-trained ConvNets fine-tuned on the 2D aggregation of STFT, MFCC, and CRP, for a fair comparison as well as to evaluate the potential of the proposed WCCGAN to generate DWT spectrograms that may also improve the performance of other classification approaches, we have evaluated the performance of GoogLeNet and AlexNet on the DWT spectrograms augmented by the proposed WCCGAN. We fine-tuned these two ConvNets with the four augmented datasets and the results are shown in Table~\ref{convnetsaug}. The results show the importance of high-level data augmentation for environmental sound classification since the performance of these two ConvNets is also improved. With respect to the values reported in Table~\ref{convnetsaug}, GoogLeNet trained on the augmented DWT spectrogram outperforms the proposed classification method on ESC-50 dataset. Moreover, the performance of these two ConvNets is very close to our classification scheme.

\begin{table}[htpb!]
\caption{Recognition accuracy of two ConvNets on the augmented DWT spectrograms of four benchmarking datasets. Value in bold indicates a better performance than those reported in Table~\ref{Rec_acc}. The 5-fold cross validation setup is applied. The mean confidence refers to the probabilities computed by the softmax layer.}
\label{convnetsaug}
\centering
\footnotesize
\renewcommand{\arraystretch}{1.3}
\begin{tabular}{|c|c|c|c|c|}
\hline
& \multicolumn{2}{c|}{Mean Accuracy} & Mean Confidence\\ \cline{2-3}
Dataset      & GoogLeNet & AlexNet & (\%) \\ \hline
ESC-10       &      0.86    &   0.85 & 78.26      \\ \hline
ESC-50       &        \textbf{0.78}   &   0.75 & 80.52      \\ \hline
UrbanSound8k &        0.93   &  0.93  &  91.02  \\ \hline
DCASE-2017   &        0.73   &  0.74 &  81.37      \\ \hline
\end{tabular}
\end{table}

Table~\ref{tab:avgrank} summarizes the comparison between all approaches with and without the proposed data augmentation through an average ranking~\cite{Brazdil2000} according to the measured mean accuracy. The proposed approach with data augmentation has the best rank among all approaches, followed by the GoogLeNet and AlexNet with data augmentation, GoogLeNet, AlexNet, and the proposed approach without data augmentation. The most impressive improvement due to the proposed data augmentation is observed for the proposed approach which moves from the last (6th) to the top spot (1st). 

\begin{table}[htpb!]
\footnotesize
\renewcommand{\arraystretch}{1.3}
\centering
\caption{Average ranking ($\bar{r}$) considering the best mean accuracy for the four datasets~\cite{Brazdil2000}.}
\begin{tabular}{|l|c|c|}
\hline
Approach & $\bar{r}$ & Rank  \\
\hline
Proposed Approach (DA) & 1.25 & 1 \\  
\hline
Proposed Approach & 5.00 & 6 \\ 
\hline
GoogLeNet (DA) & 1.50 & 2 \\  
\hline
GoogLeNet & 4.50 & 4 \\ 
\hline
AlexNet (DA) & 2.50 & 3 \\ 
\hline
AlexNet & 4.75 & 5 \\  
\hline
\end{tabular}
\label{tab:avgrank}
\end{table}

\section{Discussion}
\label{sec:disc}
We have shown the potential of the proposed WCCGAN for high-level data augmentation in improving the performance of two different supervised approaches (ConvNets and RFs). Since the proposed WCCGAN also considers inter-class and intra-class aspects to generate new samples, it allows generating more discriminating features as it improves recognition accuracy of all classifiers. Implementing low-level 1D data augmentation approaches proposed by Salamon et al.~\cite{salamon2017deep}, do not noticeably help to learn more informative features. Table~\ref{aug_comp_rev} compares the results of several low-level 1D data augmentation approaches and a single low-level 1D data augmentation approach. For instance, we augmented the environmental datasets using all 1D augmentation approaches defined in~\cite{salamon2017deep}: time-stretching with scale of 0.81, 0.93, 1.07, and 1.23; pitch-shifting with factors of 0.75, 0.9, 1.15, and 1.5 (the same parameters defined in Section \ref{sec:exp}); dynamic range compression using three parameterization from Dolby E standard and one from Icecast radio streaming server; and background noise using acoustic scenes of street-workers, street traffic, street-people, and park. Table~\ref{aug_comp_rev} shows that employing all types of low-level data augmentation do not necessarily improve the performance of the classifier.

\begin{table}[htpb!]
\caption{Comparison of 1D data augmentations approaches in terms of recognition accuracy for the proposed classification scenario in a 5-fold cross validation setup. Note that, after these 1D data augmentation, we have also augmented the DWT representations with WCCGAN.}
\footnotesize
\renewcommand{\arraystretch}{1.3}
\centering
\begin{tabular}{|c|c|c|}
\hline
& \multicolumn{2}{c|}{Mean Accuracy}\\ \cline{2-3}
Dataset & All 1D Augmentation & Pitch-shifting \\ 
\hline
ESC-10       & 0.79               & 0.87         \\ \hline
ESC-50       & 0.75               & 0.77            \\ \hline
UrbanSound8k & 0.92               & 0.94         \\ \hline
DCASE-2017   & 0.69               & 0.76         \\ \hline
\end{tabular}
\label{aug_comp_rev}
\end{table}

We have also compared the performance of the proposed WCCGAN with the CCGAN proposed by Zhu et al.~\cite{zhu2018emotion} on the DWT spectrograms. The input size of the generator and discriminator networks in the CCGAN is 48$\times$48 which is considerably smaller than our spectrogram dimensions of 768$\times$384. For a fair comparison, we have adapted the input dimensions of the networks to the size of our generated spectrograms as well as we have squeezed the DWT spectrograms to 48$\times$48 to fit them to the networks. The results of these experiments are summarized in Tables~\ref{CCGAN_RF} and~\ref{CCGAN_CNN}. Table~\ref{CCGAN_RF} shows that the proposed WCCGAN outperforms the architecture introduced in~\cite{zhu2018emotion} considering our front-end RF classifier for both input dimensions. Table~\ref{CCGAN_CNN} shows that the proposed approach also outperforms the CCGAN when we use the ConvNet proposed by Zhu et al.~\cite{zhu2018emotion} as a front-end classifier. These results show the advantage of the proposed WCCGAN and  front-end classification compared to the classification pipeline proposed in~\cite{zhu2018emotion} for spectrograms.

\begin{table}[htpb!]
\centering
\caption{Recognition accuracy of the proposed approach with different cycle-GAN augmentation architectures on DWT spectrograms. The 5-fold cross validation setup is applied and the bold values indicate the best performance.}
\label{CCGAN_RF}
\centering
\footnotesize
\renewcommand{\arraystretch}{1.3}
\begin{tabular}{|c|c|c|c|}
\hline
& \multicolumn{3}{c|}{Mean Accuracy}\\ \cline{2-4}
Dataset      & \multicolumn{2}{c|}{CCGAN~\cite{zhu2018emotion}} & WCCGAN  \\ \cline{2-4}
& (48$\times$48) & (768$\times$384) & (768$\times$384) \\ \hline
ESC-10       &      0.74    &   0.75 & \bf 0.87      \\ \hline
ESC-50       &      0.67   &   0.70 & \bf 0.77      \\ \hline
UrbanSound8k &      0.80 & 0.80 & \bf 0.94 \\ \hline
DCASE-2017   &      0.67 & 0.71 & \bf 0.76      \\ \hline
\end{tabular}
\end{table}

\begin{table}[htpb!]
\centering
\caption{Recognition accuracy of the ConvNet~\cite{zhu2018emotion} with different cycle-GAN augmentation architectures on DWT spectrograms. The 5-fold cross validation setup is applied and the bold values indicate the best performance.}
\label{CCGAN_CNN}
\footnotesize
\renewcommand{\arraystretch}{1.3}
\begin{tabular}{|c|c|c|c|c|}
\hline
& \multicolumn{4}{c|}{Mean Accuracy}\\ \cline{2-5}
Dataset      & \multicolumn{2}{c|}{CCGAN~\cite{zhu2018emotion}} & \multicolumn{2}{c|}{WCCGAN} \\ \cline{2-5}
& (48$\times$48) & (768$\times$384) & (48$\times$48) & (768$\times$384) \\ \hline
ESC-10       &      0.40    &   0.41 & 0.59 & \bf 0.67      \\ \hline
ESC-50       &      0.41   &   0.44 & 0.51 & \bf 0.59      \\ \hline
UrbanSound8k &      0.38 & 0.41 & 0.59 & \bf 0.64 \\ \hline
DCASE-2017   &      0.39 & 0.42 & 0.50 & \bf 0.52      \\ \hline
\end{tabular}
\end{table}

Finally, Table~\ref{table:meanacc} shows the mean classification accuracy of the proposed approach with and without data augmentation as well as the results obtained by other state-of-the-art classifiers described in the literature. The proposed approach achieved the highest mean accuracy for ESC-50 and DCASE-2017 and its performance is just 0.01 lower than the approach based on the decision-level fusion of two parallel ConvNets (MC-Net + LMC-Net) for the UrbanSound8k dataset. However, the best performance for the ESC-10 dataset is achieved by the Soundnet~\cite{aytar2016soundnet} which learns a multimodal representation from a very-large dataset of unlabeled videos which is further used with an SVM. Besides, the proposed approach outperforms most of the approaches trained on handcrafted features or trained on both 1D signal and spectrograms.

\begin{table*}[htpb!]
\footnotesize
\renewcommand{\arraystretch}{1.3}
\caption{Mean accuracy of different environmental sound classification approaches in UrbanSound8k, ESC-10, ESC-50 and DCASE-2017 datasets with and without data augmentation (DA). Values are rounded in two floating point precision.}
\centering 
\begin{tabular}{|l|c|c|c|c|} 
\hline 
		 & \multicolumn{4}{c|}{Mean Accuracy }\\ \cline{2-5}
Approach & UrbanSound8k & ESC-10 & ESC-50 & DCASE-2017\\ 
\hline 
\textbf{Proposed Approach (DA)} &0.94 & 0.87 & \textbf{0.77} & \textbf{0.76}\\ 
\hline
MC-Net + LMC-Net~\cite{su2019environment} & \textbf{0.95} & 0.72 & 0.74 & 0.74\\
\hline
GooLeNet and AlexNet~\cite{boddapati2017classifying} & 0.93 & 0.86 & 0.73 & NA\\
\hline
SoundNet~\cite{aytar2016soundnet}&0.79 & \textbf{0.92} & 0.74 & NA\\
\hline 
SB-ConvNets (DA)~\cite{salamon2017deep}&0.79 & 0.77 & 0.54 & 0.45\\ 
\hline
MoE~\cite{ye2017urban} & 0.77 & NA & NA & NA \\
\hline
SKM (DA)~\cite{salamon2015unsupervised}&0.76 & 0.74 & 0.56 & 0.43\\ 
\hline
SKM~\cite{salamon2015unsupervised}&0.74 & 0.71 & 0.52 & 0.36\\ 
\hline 
\textbf{Proposed Approach} & 0.73 & 0.71 & 0.55 & 0.66\\ 
\hline
PiczakConvNets~\cite{piczak2015environmental}&0.73 & 0.80 & 0.65 & 0.52\\ 
\hline
SB-ConvNets~\cite{salamon2017deep}&0.73 & 0.72 & 0.49 & 0.41\\ 
\hline 
MultiTemp~\cite{Zhu2018} & 0.72 & 0.74 & 0.71 & 0.73\\
\hline
VGG~\cite{Pons2018} & 0.70 & NA & NA & NA\\
\hline
\multicolumn{5}{l}{\scriptsize{NA: Not Available.}}
\end{tabular}
\label{table:meanacc} 
\end{table*}

\section{Conclusion}
In this paper we have shown how to structurally augment imbalanced environmental sound datasets in a high-level fashion using the proposed WCCGAN. The proposed data augmentation framework applies identity mapping to discriminator networks, which using the least-squared optimization criterion solves the gradient vanishing problem and produces nice-looking spectrograms. The importance of the high-level augmentation is more tangible for spectrograms because compared to regular computer vision datasets (e.g., ImageNet~\cite{imagenet_cvpr09}), spectrograms do not have solid objects sensitive to low-level transformations. Moreover, the total number of samples in environmental sound datasets are limited and image-to-image translation using the WCCGAN can effectively increase the size and improve the quality of the datasets. The proposed high-level data augmentation approach is also able to produce consistent samples that keep structural significance which is much more meaningful compared to other approaches such as simple image transformations or even conventional GANs. Such approaches do not allow control of the generated samples, especially regarding their structural consistency. The experimental results have shown that the WCCGAN outperforms the regular GAN since we do not have much control over consistency of the source and target inputs. Overall, high-level data augmentation using GANs translates structural components from sample to sample where low-level augmentation algorithms cannot. Furthermore, the experimental results have also shown that the WCCGAN can even improve the performance of ConvNets for the environmental sound classification task. Unfortunately, the proposed architecture for cycle-consistent GAN does not properly work in an end-to-end 1D setup. In fact, it is really costly to train and find hyperparameters for an end-to-end WCCGAN as audio waveforms have a much higher dimensionality compared to spectrograms. In spite of the high dependence of the proposed architecture on the dataset, we believe that the proposed WCCGAN can also be adapted to other datasets with some customization in the architecture of generators and discriminators and an appropriate hyperparameter tuning. The burden of hyperparameter tuning may be reduced by using a black-box optimization such as the Ortho-MAD2S~\cite{Mello2019}.

Our classification approach is a promising step towards building reliable classifiers for complex environmental sound datasets. We learn a codebook with visual words extracted by SURF detectors from augmented spectrograms organized in a unit distance to each other in a setup imposed by the $K$-Means++ algorithm. Unsupervised feature learning has shown great competence in classifying 2D representations of the environmental sound datasets. The RF classifier with 2~000 trees trained on code vectors outperformed the two ConvNets in four benchmarking datasets (ESC-10, ESC-50, UrbanSound8k, and DCASE-2017). Furthermore, besides outperforming deep models, the unsupervised feature learning approach together with the proposed architecture for the WCCGAN compares favorably with most of the current approaches for environmental sound classification. Another aspect is the reliability of the proposed approach against adversarial attacks. It is out of the scope of this paper to discuss this aspect, but it has been shown that ConvNets such as AlexNet and GoogLeNet are more vulnerable against carefully crafted adversarial examples compared to classifiers trained with SURF feature vectors~\cite{esmaeilpour2019robust}.

For the future work, in addition to improving Spherical $K$-Means++ algorithm for environmental sound classification, we would like to measure the performance of other unsupervised algorithms on the augmented DWT datasets to understand the strength of these classifiers. Besides that, we are also interested in evaluating Wasserstein GAN~\cite{arjovsky2017wasserstein} for image-to-image translation since it suffers less from oversmoothing effects. This might improve further the performance of the proposed classification approach. Finally, we would like to extend this work for structured datasets such as music datasets and evaluate the performance of the proposed data augmentation and classification approaches.   

\small
\singlespacing 
\bibliography{mybib}

\end{document}